\documentclass[letterpaper, 10 pt, conference]{ieeeconf}  

\usepackage{times}
\usepackage{epsfig}
\usepackage{graphicx}
\usepackage{amsmath}
\usepackage{amssymb}
\usepackage{algorithm}
\usepackage{hyperref}
\usepackage{verbatim}

\hypersetup{
    colorlinks=true,
    linkcolor=blue,
    filecolor=magenta,      
    urlcolor=cyan,
}

\IEEEoverridecommandlockouts                              

\title{\LARGE \bf
Scene Classification in Indoor Environments for Robots using Context Based Word Embeddings
}

\author{Bao Xin Chen, Raghavender Sahdev, Dekun Wu, Xing Zhao, Manos Papagelis and John K. Tsotsos
\thanks{All authors are with the Department of Electrical Engineering and Computer Science at York University, Toronto, Ontario, Canada
        {\tt\small \{baoxchen, sahdev, jackwu, xingzhao, papaggel, tsotsos\} @cse.yorku.ca}}%
}

\begin{document}

\maketitle
\begin{abstract}

Scene Classification has been addressed with numerous techniques in the computer vision literature. However with the increasing number of scene classes in datasets in the field, it has become difficult to achieve high accuracy in the context of robotics. In this paper, we implement an approach which combines traditional deep learning techniques with natural language processing methods to generate a word embedding based Scene Classification algorithm. 
We use the key idea that context (objects in the scene) of an image should be representative of the scene label meaning a group of objects could assist to predict the scene class.  
Objects present in the scene are represented by vectors and the images are re-classified based on the objects present in the scene to refine the initial classification by a Convolutional Neural Network (CNN). In our approach we address indoor Scene Classification task using a model trained with a reduced pre-processed version of the Places365 dataset and an empirical analysis is done on a real world dataset that we built by capturing image sequences using a GoPro camera. We also report results obtained on a subset of the Places365 dataset using our approach and additionally show a deployment of our approach on a robot operating in a real world environment. 

\end{abstract}

\section{INTRODUCTION}

Scene Classification has been extensively researched by the computer vision~\cite{zhou2017places},~\cite{xiao2010sun},~\cite{kumar2016deep} community. 
Most datasets have millions of images which makes the task challenging and thereby encourages the vision community to come up with better algorithms to achieve higher accuracies. However, on some of these datasets, current state of the art algorithms have low accuracies which renders their usage in the real world impractical. For instance the top-1 accuracy on the Places365 dataset~\cite{zhou2017places} is only around 56\%. Such an accuracy is not of much use when deploying algorithms on a robot as nearly half of the time the robot would make erroneous predictions. In the past, there have been datasets with a limited number of different places, like~\cite{sahdev2016indoor},~\cite{lazebnik2006beyond}, which had only upto 17 places, however these datasets' usage has been rendered obsolete due to already achieved near human accuracy ($\approx$95\%). 

In this paper, we propose an approach for Scene Classification in real time in the context of robotics. We primarily target indoor scenes in this paper. 
In indoor environments, GPS cannot be relied on for metric level precision. We can only get a rough estimate of the position of the robot from GPS in indoor scenes. We leverage this inaccuracy in the GPS to reduce the search space of the places. We propose a taxonomy based approach to perform Scene Classification by dividing the places according to the area type (e.g., \textit{school, shopping-mall, home, etc}.). Once we know roughly which region the robot is in, we need to only search for a subset of places rather than searching all the classes. For example, from GPS we can know that the robot is in a school environment. This would reduce the search space to places like \textit{corridors, lecture hall, seminar room, washroom, etc}. (see table~\ref{dataset_testing_labels}). We no longer need to consider the places that belong to any unrelated class like \textit{living room, pet shop, jewelry store, forest, waterfall, mountains}, etc. After this step, many classes are pruned leaving a smaller set of candidate classes to select from. 
To classify the places, we propose to use different CNN models for each of the indoor areas present in the Places365 dataset~\cite{zhou2017places}.

\begin{figure}[t]
\begin{center}
\includegraphics[width=3in]{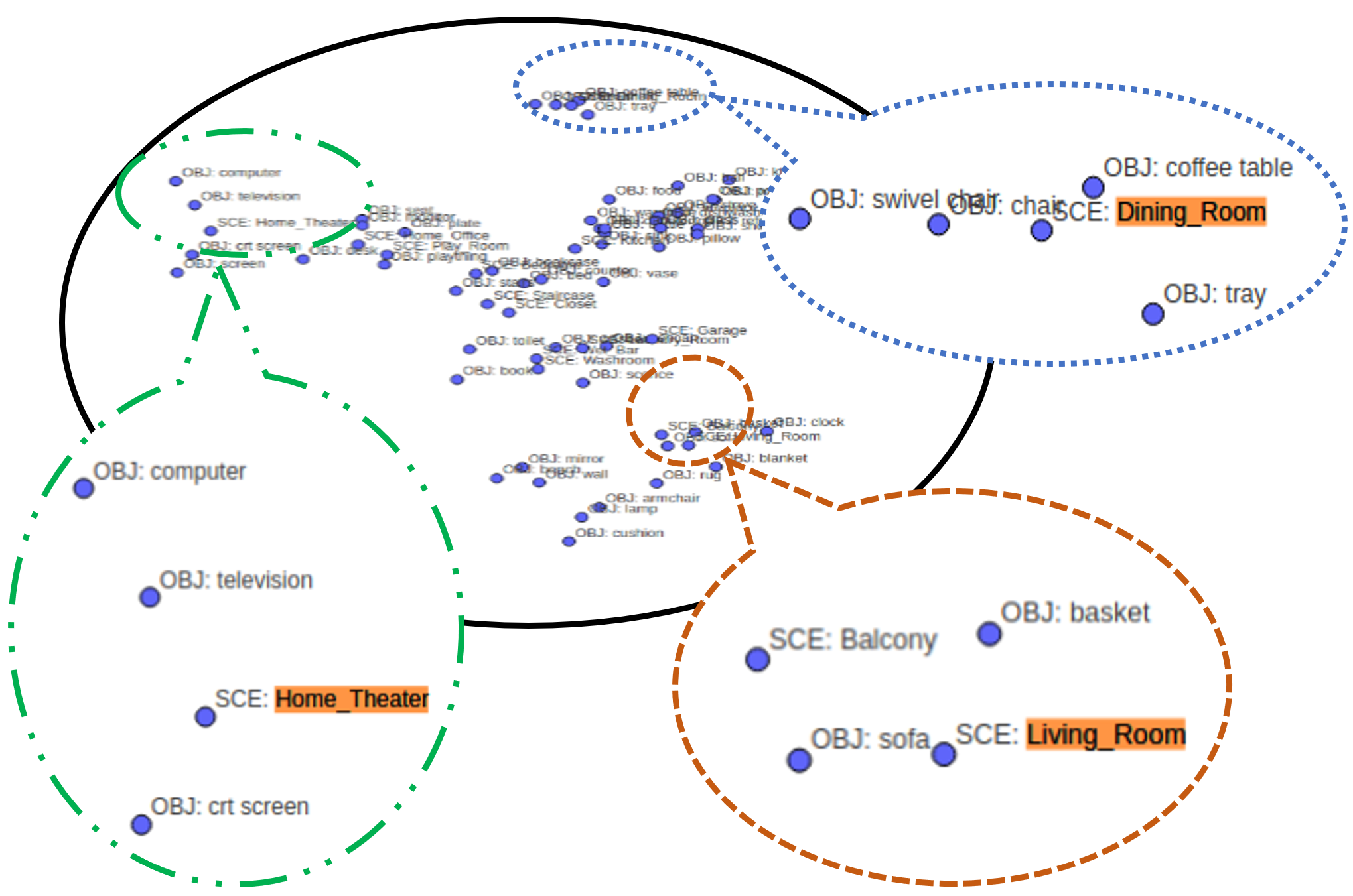}
\end{center}
\caption[]{Object and Scene vectors represented in a 2D space after learning from our word2vec model. Objects related to specific scenes are closer to each other, e.g., \textit{coffee table, tray} in \textit{Living Room}; \textit{crt screen, television} in \textit{Home Theatre}. Figure generated using tSNE Visualization tool~\footnotemark}
\label{fig:vectors}

\end{figure}
\footnotetext{{\url{https://cs.stanford.edu/people/karpathy/tsnejs/csvdemo.html}}}


\begin{figure*}[t]
\begin{center}
\includegraphics[width=5.5in]{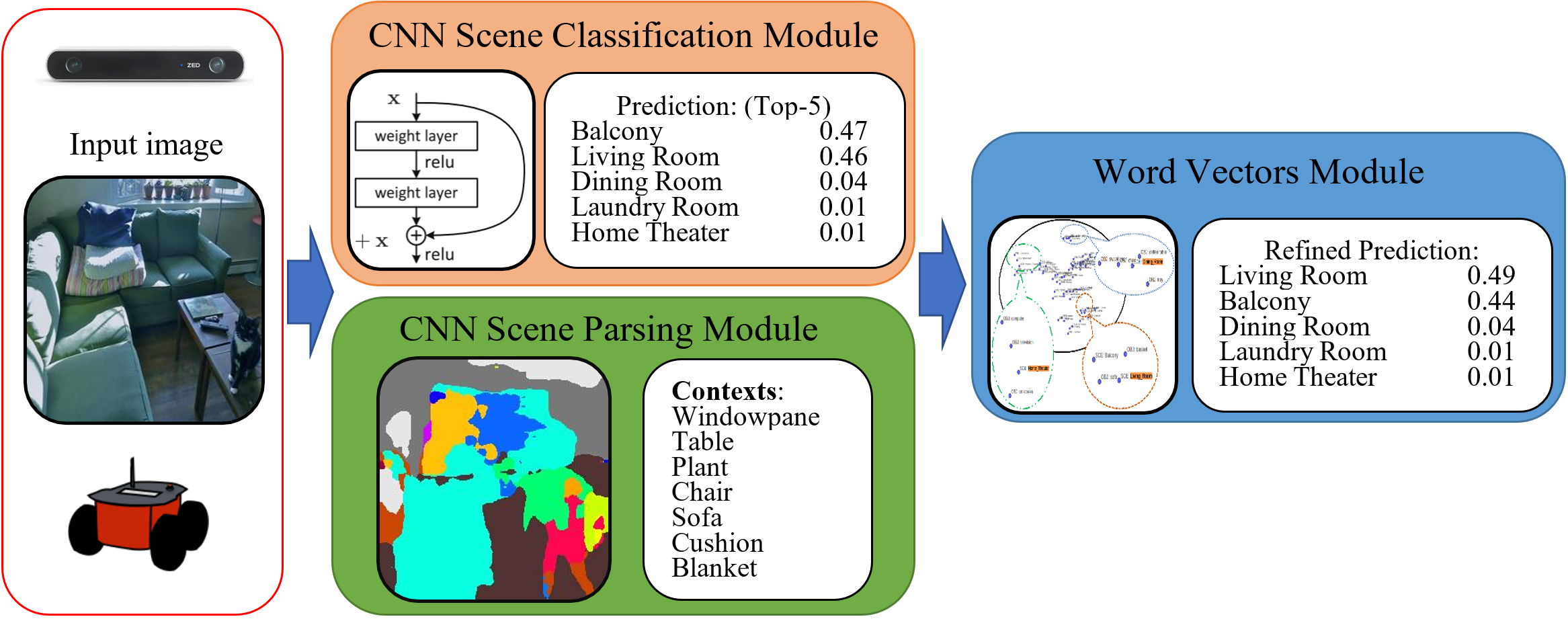}
\end{center}
\caption{Overview of our algorithm. Our approach takes an RGB image as input. Then passes it onto two CNN modules. The top module is for Scene Classification, which gives us an initial top-5 prediction. The bottom module is for Scene Parsing, which detects the scene contents including background (\textit{window pane, plant, etc.}) and foreground (\textit{table, chair, sofa, etc.}) objects. The Word Vectors Module computes a vector for the contents of the image, and a vector for each of the top-5 predicted labels. Then Word Vectors Module refines the ranking of these top-5 predicted labels by comparing the vector similarity.}
\label{fig:system_overview}
\end{figure*}

After employing a CNN for Scene Classification, we obtain the top-5 predictions for a given image. To further refine the accuracy of the prediction, we make use of different objects present in the scene which are detected using a Scene Parsing Module (see Figure~\ref{fig:system_overview}). Each object present in the image is represented using a word-embedding~\cite{mikolov2013efficient}. In most cases, these object embeddings are similar for objects belonging to a particular scene and the scene embedding itself, e.g., \textit{tray, coffee table, chairs} in a \textit{dinning room} (See Figure~\ref{fig:vectors}) should have similar embedding. Using these embeddings, we further refine the top-5 scores to improve the accuracy. 
Finally, we build our own test dataset to validate our approach as there are many inconsistencies in the Places365 dataset as described in section~\ref{dataset}. We also report the performance on a subset of the Places365 dataset.

The major contributions of this work are: $(i)$ A taxonomy based approach to make Scene Classification work in real time with high accuracy for robots using GPS information, $(ii)$ An empirical evaluation showing performance of our proposed approach, $(iii)$ A real world dataset for the task of indoor Scene Classification.
The paper is structured as follows: we describe the relevant work in Section~\ref{relatedwork}. Section~\ref{approach} describes our proposed approach. Section~\ref{dataset} describes our test dataset. Experimental results of our approach are described in Section~\ref{evaluation} and finally we conclude our work in Section~\ref{conclusion}.




\section{Related Work}
\label{relatedwork}
In this section, we give a brief overview of the existing approaches used for Scene Classification. 
Li et al.~\cite{li2010object} used object detectors as features to form an Object Bank Representation to assist Scene Classification. Our proposed approach is similar to~\cite{li2010object}, instead of using an Object Bank representaion, we employed a word vector~\cite{mikolov2013efficient} based feature representation to find similarities between objects in the image and the scene category. Yang et al.~\cite{yang2007evaluating} used key point detection to create feature vectors. Bosch et al.~\cite{bosch2008scene} used visual vocabulary as features to train a Support Vector Machine (SVM) classifier. 
Furthermore, different types of CNN architectures (AlexNet~\cite{krizhevsky2012imagenet}, VGG~\cite{simonyan2014very}, GoogleNet~\cite{szegedy2015going}) were used in Scene Classification recently. 
These CNNs achieved decent state-of-the-art performance ($\approx$56\%) on Places365~\cite{zhou2017places} dataset. 
As well as, Khan et al.~\cite{khan2017scene} integrated Places-VGG with Spectral Features to improve Scene Classification. However, the datasets they used are not appropriate for real world application. 


The CNN generally learns a feature representation of an image but the network can not tell us the object relations in the image. However, a special CNN architecture (SegNet~\cite{badrinarayanan2015segnet})  can be used to perform pixel level segmentation (Scene Parsing). A newly released pre-trained CNN model~\cite{zhou2017scene} based on a cascade segmentation module can find 150 types of contents in an image, which includes background (\textit{wall, doors, windows, floor, etc.}) and foreground (\textit{person, television, table, chair, etc.}) information. There are several approaches which can detect objects in an image; YOLO~\cite{redmon2016yolo9000} detects most of the foreground objects. But their approach does not yield anything about the background of the image. Background information also plays an important role when a human is classifying a scene. 

Word vectors \cite{pennington2014glove} are very popular in Natural Language Processing (NLP). This is also known as word embedding. It improves the performance of LSTM-CRF \cite{huang2015bidirectional} significantly on the named entity recognition systems without any language-specific knowledge \cite{lample2016neural}. Word vectors were trained to learn the vector representation of a given vocabulary. Word vectors have been used in many domains like sentiment analysis \cite{mikolov2013efficient}, detecting magnitude of events \cite{agrawal2016detecting}, named entity recognition ~\cite{xu2017local}, etc. 


Scene Classification can also be useful for Place Recognition as it reduces the search space. Alternatively Place recognition can also be used to tackle scene classification (in case where all test images come from a set of predefined paces and each place is associated with a scene type). Some of the existing Place recognition works include~\cite{kumar2016deep},~\cite{hou2017bocnf},~\cite{sahdev2016indoor},~\cite{shakeri2016illumination}.



\textbf{Data sets:} Many well-known data sets and benchmarks exist. For instance, Places88 \cite{zhou2014learning} is the very first version of MIT Places benchmark\footnote{http://places.csail.mit.edu/index.html}. It has 88 scene categories. The latest Scene Classification data set is MIT Place365 \cite{zhou2017places}, which has 365 scene categories. In this paper we use Place365 as our training and evaluation data-set, as it is the largest scene classification data-set, and it contains broad categories. Some other old datasets also exist for Scene Classification. The Pascal VOC \cite{everingham2010pascal} data set has scene context annotations which are used for object detection and object segmentation tasks. Another data set is MS-COCO \cite{lin2014microsoft}. COCO is mainly used for instance detection and scene segmentation. The pre-trained CNN Scene Parsing model \cite{zhou2017scene} used in this paper is trained on COCO.

\section{Approach}
\label{approach}

Our aim is to provide a Scene Classification approach which can perform well in a known real world environment (e.g., \textit{school, home, shopping mall}). We build a taxonomy of different environments each containing separate scenes as shown in Table ~\ref{dataset_testing_labels}. We pre-process the Places365 dataset to clean and reduce it to have only sensible indoor places for our use. Places365 has many inconsistencies as described in section ~\ref{dataset}. 

We use two existing CNN models, one for Scene Classification ~\cite{zhou2017places} and the other for Scene Parsing ~\cite{zhou2017scene}. Figure \ref{fig:system_overview} shows the overview of our approach.
The input is an RGB image for the two CNN modules. We call them CNN Scene Classification Module and CNN Scene Parsing Module (Figure \ref{fig:system_overview}). The Classification Module computes the initial raw top-5 predictions. The Scene Parsing module computes the scene contents from foreground and background. The Word Vector module computes the vector similarity between the contents present in the input image and the top-5 predicted labels. Using the computed similarity score, we output a refined re-ranked top-5 labels for the input image. We describe the details about each module in following subsections.

\subsection{CNN Scene Classification Module}

In this module, we train the ResNet models (one for each environment) on a reduced version of the Places365 dataset (based on our taxonomy, see table ~\ref{dataset_testing_labels}). 
The code and model parameters can be downloaded from this link~\footnote{\url{http://places2.csail.mit.edu/}}. CNN Scene Classification Module computes the top-5 prediction labels which is further used by the Word Vectors Module to refine these prediction scores. 


\subsection{CNN Scene Parsing Module}

In this module, we use a pre-trained CNN Scene Parsing model~\footnote{\url{https://github.com/CSAILVision/sceneparsing}}. The CNN model was trained on ADE20K dataset \cite{zhou2017scene}. It is trained to detect 150 different object categories 
from the given input image. The Scene Parsing module tells us the different objects/contents present in a given input image. For a kitchen input image, the parser would output the presence of \textit{kettle, stove, oven, glasses, plates, etc}. We convert these object labels as English words and pass that onto Word Vectors module, described in the section below.

\subsection{Word Vectors Module}
 
Now we know the objects present in each of the images in the dataset. Knowing the top-5 predictions of each image and the objects present in each of the images, we need to make use of this information to refine the top-5 scores.

This section is the main contribution of this work. First, we define some of the notations being used and then describe our approach using these notations.

\textbf{Notations used:} let $V$ be the dictionary/vocabulary containing the objects, $O$ and the scenes, $S$ present in the dataset. $V = \{O, S\}$, 
where $O = \{{o_{1},o_{2},o_{3},o_{4}......o_{150}}\}$ and $S=\{{s_{1},s_{2},s_{3}...s_{n}}\}$, where each $o_i$ is a vector representing each of the objects and each $s_{j}$ is a vector representing the scene classes present in the dataset. 
We further define a 2D weight matrix, $W$ of scalars. The dimensionality of the weight matrix is $|O|$ * $|S|$. where $|O| = 150$, the number of different objects we trained on and $|S| = n$, the number of defined scenes in a particular environment (e.g., for school, $n=24$). The matrix $W =\{ {w_{1,1},..w_{i,j},..w_{150,n}}\}$ where $i \in{[1,150]}$ and $j \in{[1,n]}$. Also we define $T$ as the matrix containing the top-5 predictions for each of the images. $T$ has dimensionality $5$ * $|D|$ where $|D|$ is the cardinality of the dataset. $T_{k}$ represents the row of the top-$5$ predicted classes in the $k^{th}$ Image, $I_{k}$. Similar to $T$, $ACC$ is the confidence for the predicted top-5 classes.  $IO_{k}$ is the set of objects present in the $k^{th}$ image, $I_{k}$. A sample $IO_{k}$ (for image, $I_k$) would be like $\{{o_{14},o_{52},o_{78},o_{113},o_{143}, o_{149}}\}$ implying say an input image (e.g., \textit{home\_office}) consists of 6 objects \textit{\{laptop, keyboard, table, desk, chair, book\}}. 


\begin{algorithm}
\label{algorithm_word_vec}
\texttt{\textbf{\small{}{}{}Input:}}{\small \par}

{\small{}{}{}A dataset, $D$ of RGB images, $I_{k}\in\mathbb{D}$: $D=\left\{ I_{1},I_{2},...,I_{n}\right\} $}{\small \par}

{\small{}{}{}A vocabulary, $V$ of object vectors, $O$ and $S$ : $V=\left\{ O,S\right\} $}{\small \par}

{\small{}{}{}A list of Object vectors, $O$ in $V :$ $ O=\left\{ o_{1},o_{2},...,o_{150}\right\} $}{\small \par}

{\small{}{}{}A list of Scene vectors, $S$ in $V :$ $ S=\left\{ s_{1},s_{2},...,s_{n}\right\} $}{\small \par}


{\small{}{}{}A weight matrix, $W$ for objects and scene, $W : \{w_{i,j},$ where $i\in[1,150];j\in[1,n] \}$}{\small \par}

{\small{}{}{}\medskip{}
 }{\small \par}

\texttt{\textbf{\small{}{}{}Output:}}{\small \par}


{\small{}{}{}Refined Top5 Prediction, $RT_{k}$ for each image, $I_{k}$ after Refinement }{\small \par}

{\small{}{}{}\medskip{}
 }{\small \par}

\texttt{\textbf{\small{}{}{}Procedure 1, main:}}{\small \par}

{\scriptsize{}{}{}1.}\textbf{\small{}{}{}\ for}{\small{}{}{}
$I_{k}\in D$ }\textbf{\small{}{}{}do}{\small \par}

{\scriptsize{}{}{}2.}{\small{}{}{}\qquad{}$T_{k}$, $ACC_{k}$ = 
$Scene Classification Module\left(I_{k}\right)$}{\small \par}

{\scriptsize{}{}{}3.}{\small{}{}{}\qquad{}$IO_{k}$ =  
$Scene Parsing Module\left(I_{k}\right)$}{\small \par}

{\scriptsize{}{}{}4.}{\small{}{}{}\qquad{}$RT_{k}$ = 
$Word Vector Module\left(IO_{k},T_{k},ACC_{k}\right)$}{\small \par}



{\scriptsize{}{}{}5.}{\small{}{}{}}\textbf{\small{}{}\ return}{\small{}{}{}
$RT$}{\small \par}

{\small{}{}{}\medskip{}
 }{\small \par}
 
\texttt{\textbf{\small{}{}{}Procedure 2, WordVectorModule ($IO_{k},T_{k},ACC_{k}$) :}}{\small \par}



{\scriptsize{}{}{}1.}{\small{}{}{} $Similarity[5]$ = 
$\{0,0,0,0,0\}$}{\small \par}

{\scriptsize{}{}{}2.}\textbf{\small{}{}{}\ for}{\small{}{}{}
$s_{j}\in T_{k}$ }\textbf{\small{}{}{}do}{\small \par}

{\scriptsize{}{}{}3.}{\small{}{}{}\qquad{} $IVector_{k}(j)$ = 
$<0> vector$}{\small \par}

{\scriptsize{}{}{}4.}\textbf{\small{}{}{}\qquad{}\ for}{\small{}{}{}
$o_{i}\in IO_{k}$ }\textbf{\small{}{}{}do}{\small \par}

{\scriptsize{}{}{}5.}{\small{}{}{}\qquad{}\qquad{} $IVector_{k}(j)\ += o_{i}*w_{i,j}$  
}{\small \par}

{\scriptsize{}{}{}6.}{\small{}{}{}\qquad{} $Similarity[j]$ = 
$cosine(IVector_{k}(J), s_{j})$}{\small \par}

{\scriptsize{}{}{}7.}{\small{}{}{} $Similarity = Normalize(Similarity)$
}{\small \par}
{\scriptsize{}{}{}8.}{\small{}{}{} $RT_{k} = descendingOrder(ACC_{k}*Similarity)$
}{\small \par}

{\scriptsize{}{}{}9.}{\small{}{}{}}\textbf{\small{}{}\ return}{\small{}{}{}
$RT_k$}{\small \par}

\protect\protect\caption{\label{alg:Pseudocode-of-the}Pseudocode of the proposed approach}
\end{algorithm}

Now we have the initial top-5 predicted classes $T_{k}$ for each image, $I_{k}$ obtained from the Scene Classification module and the set of objects, $IO_{k}$ present in each image obtained from the scene parser module. Our approach can be summarized in Algorithm 1. 
To compute the refined top-5 scene classes predictions, $RT_k$: first we compute a vector representation for the image with respect to each of the top-5 predicted classes. We compute a weighted vector sum of the detected objects in the image to represent the image vector. The vector representation for a given input image, $I$ with respect to a scene $s_{j}$ is computed as follows: 

\begin{equation}
IVector_{k}(j)= \sum_i w_{i,j}*o_{i}
\end{equation}

\begin{figure}[t]
\begin{center}
\includegraphics[width=3.2in]{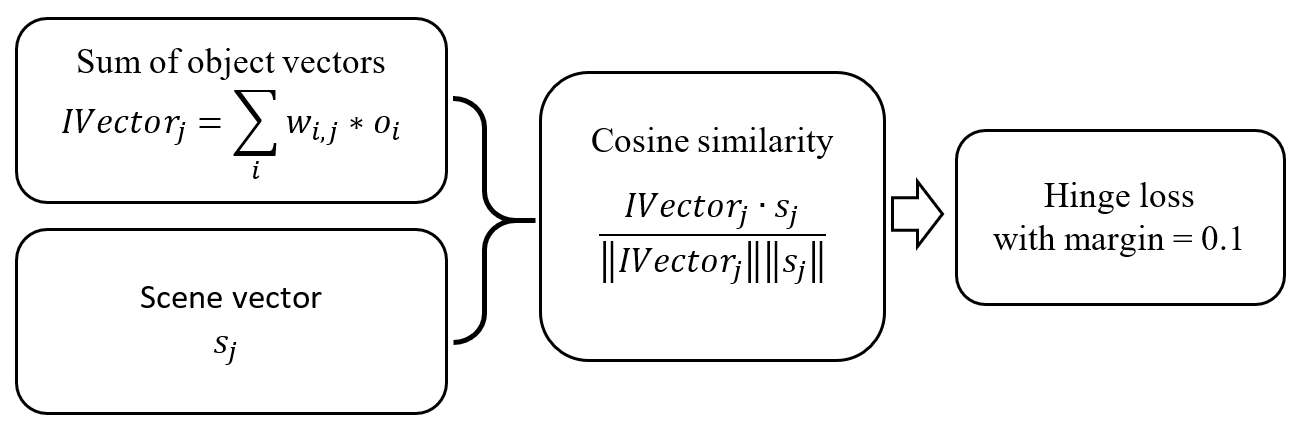}
\end{center}
\caption{Training Model for learning the new word2vector $V$, and the weighted table $W$.}
\label{fig:trainingModel}
\end{figure}

Here $o_{i}\in IO_{k}$. The weights, $w_{i,j}$ and vectors, $o_{i}$ are learned by the word vector module (see Figure~\ref{fig:trainingModel}). $IVector_{k} (j)$ is the vector representation of the input image, $I_{k}$ in terms of the objects present \textit{w.r.t.} to scene label $j$. Refer to line 4,5 in Algorithm 1, Procedure 2.

\textbf{Learning the object and scene vectors, $V$, and the weight matrix, $W$:}


Vectors in $V$ are initialized by the pre-trained word2vector~\footnote{\url{https://radimrehurek.com/gensim/models/word2vec.html}}, and $W$ are initialized by \textit{tf–-idf}~\footnote{\url{http://www.tfidf.com/}}. When computing \textit{tf-idf}, we considered each scene class is a document, and the objects in the image are the words. The object representing the scene has a higher weight and the object occurring less frequently in the scene would have a lower weight. We pre-compute the training predictions, $T$, $ACC$ and objects in images, $IO$. We feed this information into our cosine similarity training model (see Figure \ref{fig:trainingModel}) to learn the weights, $W$ and Vocabualry, $V$. 
The loss function used in our model was a hinge loss with margin equal to $0.1$. 

There are some objects, $o_{i} \in O$ which might exist in most of the scenes. Such objects contribute less in determining the class of the scene. So a lesser weight should be assigned to such objects while computing the weighted sum representation of the image vector, $IVector_{k}(j)$. On the other hand, if an object exists only in some particular scenes, we need to increase the weight of this object. This is intuitive, for instance: the chances of finding a \textit{``sofa''} is more likely in a \textit{``living-room''} scene than any other place. Now we describe the dataset we built to validate our approach and provide empirical results obtained on our dataset and the Places365 dataset.

\section{The Dataset}
\label{dataset}

One of the current widely used datasets for Scene Classification is the Places365 dataset~\cite{zhou2017places}. This dataset is the latest and largest Scene Classification dataset with more than 1 million images. Places365 has 365 scene categories with 3000 to 5000 images per class. 
However in this dataset there are many confusing places that are practically the same but have been labelled distinctly. For example, the places \textit{\{bedroom, dorm room\}, \{lecture room, classroom\}, \{pharmacy, drug-store\},} etc. are some sets of places that are referring to the same but labelled differently. Moreover, after a close inspection of this dataset, it was found that many images in the dataset are incorrectly labelled. Due to these mentioned issues with the Places365 dataset, we decided to merge certain classes into one before training our models (see taxonomy file on the project page$^{~\ref{project_page_link}}$). Additionally since in our work, we are only concerned about indoor places, we remove all outdoor places in the Places365 dataset. After this pre-processing step of removal and merging of places, we get a total of 156 different scenes for 8 different environments. We train our model using this pre-processed dataset.

For testing our proposed approach, we built a dataset in the real world of 69 different places (see table~\ref{dataset_testing_labels}), which we arranged in a hierarchical manner with 2 levels. First level (environment-type) lists out the major areas like school, home, shopping mall, etc. Each of the categories in the first level has sub-class scenes associated with them. For e.g., a shopping mall would have a set of places like \textit{gift-shop, food-court, salon, drug-store, etc}. A detailed list of places in our testing dataset can be found in table~\ref{dataset_testing_labels} and at our project page~\footnote{\url{http://jtl.lassonde.yorku.ca/2018/04/scene-classification-robots/}\label{project_page_link}}. We have a total of 10,000 testing images in our dataset. Out of these approximately 1000 images were extracted from videos on youtube as some places were not easily available in the real world and others were captured from a GoPro camera.

\begin{table*}
\centering
\caption{Sample data classes in our Dataset. For complete taxonomy see project page$^{~\ref{project_page_link}}$}
\label{dataset_testing_labels}
\begin{tabular}{|p{2cm}|p{14.9cm}|}
\hline
\textbf{First Level Place (Environments)} & \textbf{Second Level Place (Scenes)}
\\ \hline
\textbf{School \newline(24-scenes)}            & Classroom, Kindergarten, Office, MeetingRoom, ComputerLab, ChemistryLab, BiologyLab, PhysicsLab, Library, Corridor, Elevator, Escalator, Cafeteria, Washroom, Auditorium, Gymnasium, LockerRoom, IndoorSwimmingPool, BasketballCourt, VolleyballCourt, BadmintonCourt, TableTennis, DormRoom, Lobby
\\ \hline
\textbf{Home \newline(14-scenes)}              & Kitchen, WetBar, LivingRoom, DiningRoom, Bedroom, Closet, PlayRoom, HomeTheater, HomeOffice, LaundryRoom, Washroom, Garage, Staircase, Balcony
\\ \hline
\textbf{Shopping Mall (31-scenes)}     & BookStore, CandyStore, VideoStore, MusicStore, HardwareStore, ShoeShop, DrugStore, ToyStore, ClothingStore, HatShop, FloristStore, JewelryStore, Optician, Supermarket, Bakery, Salon, PetShop, GiftShop, Foodcourt, Bar, Restaurant, CoffeeShop, TeaShop, DepartmentStore, Reception, Fountain, Elevator, Escalator, Washroom, IndoorParking, MovieTheater
\\ \hline
\end{tabular}
\end{table*}

\begin{table*}[t]
  \begin{center}
    \caption{Top-1 accuracy on the reduced validation set on Places365 and our test set.}
    \label{tab:table1}
    \begin{tabular}{l|c|c|c|c|c|c}
       {} & \multicolumn{2}{c}{School (24)} & \multicolumn{2}{c}{Home (14)} & \multicolumn{2}{c}{Shopping Mall (31)} \\
       {} & Places365 val.-set & our test-set & Places365 val.-set & our test-set & Places365 val.-set & our test-set \\
      \hline
      Places365-ResNet &  51.45\% & 58.18\% &  57.13\% & 60.68\% &  50.17\% & 62.58\%\\
      ResNet50 &  73.82\% & 90.33\% &  83.46\% & 92.03\% &  \textbf{70.47\%} & 87.31\%\\
      ResNet50+Word2Vec (Ours) & \textbf{74.28\%} & \textbf{92.25\%} & \textbf{83.67\%} & \textbf{93.27\%} & 70.44\% & \textbf{87.39\%}\\
    \end{tabular}
    
  \end{center}
\end{table*}


\section{Evaluation}
\label{evaluation}
In our taxonomy, we described 8 different environment types (\textit{School, Shopping Mall, Home, Condo Buildings, Airport, Public Transit, Hospital, and Hotel}) using 156 indoor scenes. In this work, we select 3 environments to evaluate our approach, which are \textit{School, Home}, and \textit{Shopping Mall} (see Table~\ref{dataset_testing_labels}).

In our experiments, we trained a ResNet model for each environment with the selected scenes. The performance is listed in Table~\ref{tab:table1}. As one would expect, the ResNet model performs well on the reduced Places365 validation set. The performance on our test set is even higher, since our dataset is labeled with fewer human errors. Furthermore, we used a GoPro camera with wide-angle view mode, so the images perfectly represent the scenes. Overall, our method is better than a CNN model (ResNet50) in terms of accuracy. Our results proved that the \textit{``image context (objects)''} plays a vital role in Scene Classification. Although, the refinement on Shopping Mall scenes is not noticeable, this is because the CNN Scene Parsing model we used could not find objects that can distinguish the scenes well enough. For instance, the model could not detect shoes, watches, hats, etc., which are commonly found in shopping malls.

\begin{figure*}[t]
\begin{center}
\includegraphics[width=5.8in]{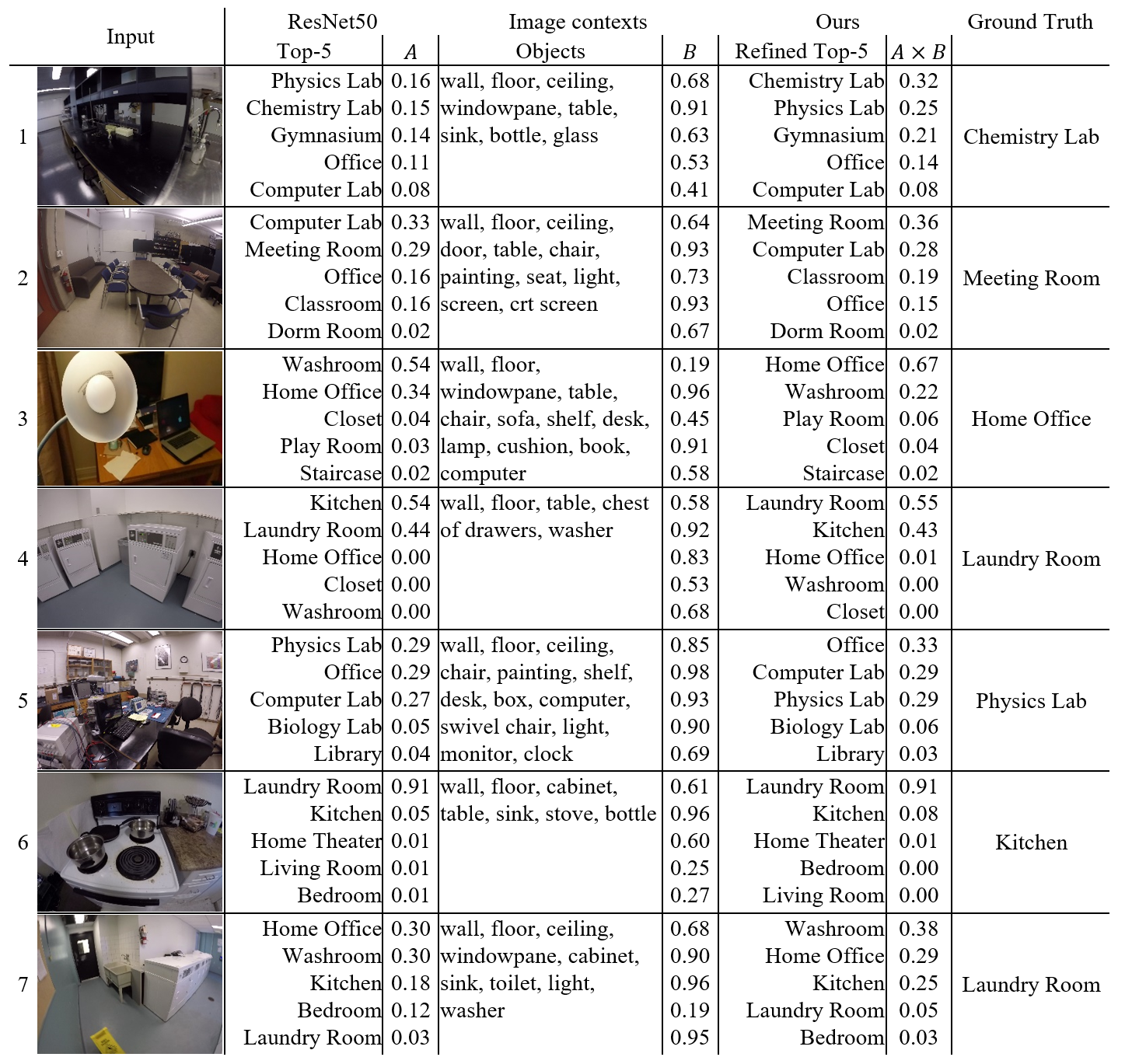}
\end{center}
\caption{Column A is the top-5 confidences obtained from the CNN Scene Classification module. Column B is the cosine similarity score from the Word Vectors module, Column AxB represents the refined top-5 predictions (refer to line 8 in Algorithm 1, Procedure 2). Sample results: (1)-(4) are correct examples, (5)-(7) are incorrect examples on our test set. The first 4 rows show that our model correctly predicted the scene labels. In the first example, CNN gave an unclear result by predicting the image as \textit{physics lab, chemistry lab, gymnasium} with very similar scores. But, our word2vec model gave a very high score on \textit{chemistry lab}. By combining the two scores, our model gave a higher predicted score on the ground truth label. On the other hand, our method also yielded some bad predictions. For row (5), our word2vec model provided a very high confidence on the false label \textit{office} by using the context of the image. As a result, this image was incorrectly classifying as \textit{office} instead of \textit{physics lab}. Another example in row (6), the CNN model very confidently predicted a \textit{kitchen} as \textit{laundry room}. Even though word2vec said it is a \textit{kitchen}, our model still could not refine the CNN predictions.  }
\label{fig:results}
\end{figure*}

Figure~\ref{fig:results} discusses the results of some images in our dataset. Finally, we also demonstrate our approach by deploying our algorithm on a Pioneer3AT robot in a university environment. The robot was configured with a ZED camera from which monocular images were used by our algorithm. The robot was equipped with a Razer Blade laptop with a GTX 1060 GPU. The demo video of the deployment is available on our project web-page$^{~\ref{project_page_link}}$.

\section{Conclusion}
\label{conclusion}

In this paper, we introduced a new method for Scene Classification using a taxonomy for different indoor environments. Using our approach, robots can recognize different indoor places with high confidence/accuracy. A hierarchical taxonomy of places allowed us to prune out many irrelevant classes, thereby reducing the complexity/training time of our approach.
A word embedding based approach was implemented to refine the top-5 scores for the scenes. It was shown that context has the potential to improve the Scene Classification results to some extent. 
We additionally tested our approach with a real world dataset that we built to show the practical applicability of our approach. We also deployed our algorithm on a robot in a university environment. On our dataset we could get promising results for doing Scene Classification for robots.



{\small
\bibliographystyle{IEEEtran.bst}
\bibliography{main}

\begin{thebibliography}{10}
\providecommand{\url}[1]{#1}
\csname url@rmstyle\endcsname
\providecommand{\newblock}{\relax}
\providecommand{\bibinfo}[2]{#2}
\providecommand\BIBentrySTDinterwordspacing{\spaceskip=0pt\relax}
\providecommand\BIBentryALTinterwordstretchfactor{4}
\providecommand\BIBentryALTinterwordspacing{\spaceskip=\fontdimen2\font plus
\BIBentryALTinterwordstretchfactor\fontdimen3\font minus
  \fontdimen4\font\relax}
\providecommand\BIBforeignlanguage[2]{{%
\expandafter\ifx\csname l@#1\endcsname\relax
\typeout{** WARNING: IEEEtran.bst: No hyphenation pattern has been}%
\typeout{** loaded for the language `#1'. Using the pattern for}%
\typeout{** the default language instead.}%
\else
\language=\csname l@#1\endcsname
\fi
#2}}

\bibitem{zhou2017places}
B.~Zhou, A.~Lapedriza, A.~Khosla, A.~Oliva, and A.~Torralba, ``Places: A 10
  million image database for scene recognition,'' \emph{IEEE Transactions on
  Pattern Analysis and Machine Intelligence}, 2017.

\bibitem{xiao2010sun}
J.~Xiao, J.~Hays, K.~A. Ehinger, A.~Oliva, and A.~Torralba, ``Sun database:
  Large-scale scene recognition from abbey to zoo,'' in \emph{Computer vision
  and pattern recognition (CVPR), 2010 IEEE conference on}.\hskip 1em plus
  0.5em minus 0.4em\relax IEEE, 2010, pp. 3485--3492.

\bibitem{kumar2016deep}
D.~Kumar, ``Deep learning based place recognition for challenging
  environments,'' Master's thesis, University of Waterloo, 2016.

\bibitem{sahdev2016indoor}
R.~Sahdev and J.~K. Tsotsos, ``Indoor place recognition system for localization
  of mobile robots,'' in \emph{Computer and Robot Vision (CRV), 2016 13th
  Conference on}.\hskip 1em plus 0.5em minus 0.4em\relax IEEE, 2016, pp.
  53--60.

\bibitem{lazebnik2006beyond}
S.~Lazebnik, C.~Schmid, and J.~Ponce, ``Beyond bags of features: Spatial
  pyramid matching for recognizing natural scene categories,'' in
  \emph{Computer vision and pattern recognition, 2006 IEEE computer society
  conference on}, vol.~2.\hskip 1em plus 0.5em minus 0.4em\relax IEEE, 2006,
  pp. 2169--2178.

\bibitem{mikolov2013efficient}
T.~Mikolov, K.~Chen, G.~Corrado, and J.~Dean, ``Efficient estimation of word
  representations in vector space,'' \emph{arXiv preprint arXiv:1301.3781},
  2013.

\bibitem{li2010object}
L.-J. Li, H.~Su, L.~Fei-Fei, and E.~P. Xing, ``Object bank: A high-level image
  representation for scene classification \& semantic feature sparsification,''
  in \emph{Advances in neural information processing systems}, 2010, pp.
  1378--1386.

\bibitem{yang2007evaluating}
J.~Yang, Y.-G. Jiang, A.~G. Hauptmann, and C.-W. Ngo, ``Evaluating
  bag-of-visual-words representations in scene classification,'' in
  \emph{Proceedings of the international workshop on Workshop on multimedia
  information retrieval}.\hskip 1em plus 0.5em minus 0.4em\relax ACM, 2007, pp.
  197--206.

\bibitem{bosch2008scene}
A.~Bosch, A.~Zisserman, and X.~Mu{\~n}oz, ``Scene classification using a hybrid
  generative/discriminative approach,'' \emph{IEEE transactions on pattern
  analysis and machine intelligence}, vol.~30, no.~4, pp. 712--727, 2008.

\bibitem{krizhevsky2012imagenet}
A.~Krizhevsky, I.~Sutskever, and G.~E. Hinton, ``Imagenet classification with
  deep convolutional neural networks,'' in \emph{Advances in neural information
  processing systems}, 2012, pp. 1097--1105.

\bibitem{simonyan2014very}
K.~Simonyan and A.~Zisserman, ``Very deep convolutional networks for
  large-scale image recognition,'' \emph{arXiv preprint arXiv:1409.1556}, 2014.

\bibitem{szegedy2015going}
C.~Szegedy, W.~Liu, Y.~Jia, P.~Sermanet, S.~Reed, D.~Anguelov, D.~Erhan,
  V.~Vanhoucke, and A.~Rabinovich, ``Going deeper with convolutions,'' in
  \emph{Proceedings of the IEEE conference on computer vision and pattern
  recognition}, 2015, pp. 1--9.

\bibitem{khan2017scene}
S.~H. Khan, M.~Hayat, and F.~Porikli, ``Scene categorization with spectral
  features,'' in \emph{Proceedings of the IEEE Conference on Computer Vision
  and Pattern Recognition}, 2017, pp. 5638--5648.

\bibitem{badrinarayanan2015segnet}
V.~Badrinarayanan, A.~Kendall, and R.~Cipolla, ``Segnet: A deep convolutional
  encoder-decoder architecture for image segmentation,'' \emph{IEEE
  Transactions on Pattern Analysis and Machine Intelligence}, 2017.

\bibitem{zhou2017scene}
B.~Zhou, H.~Zhao, X.~Puig, S.~Fidler, A.~Barriuso, and A.~Torralba, ``Scene
  parsing through ade20k dataset,'' in \emph{Proceedings of the IEEE Conference
  on Computer Vision and Pattern Recognition}, 2017.

\bibitem{redmon2016yolo9000}
J.~Redmon and A.~Farhadi, ``Yolo9000: better, faster, stronger,'' \emph{arXiv
  preprint arXiv:1612.08242}, 2016.

\bibitem{pennington2014glove}
J.~Pennington, R.~Socher, and C.~Manning, ``Glove: Global vectors for word
  representation,'' in \emph{Proceedings of the 2014 conference on empirical
  methods in natural language processing (EMNLP)}, 2014, pp. 1532--1543.

\bibitem{huang2015bidirectional}
Z.~Huang, W.~Xu, and K.~Yu, ``Bidirectional lstm-crf models for sequence
  tagging,'' \emph{arXiv preprint arXiv:1508.01991}, 2015.

\bibitem{lample2016neural}
G.~Lample, M.~Ballesteros, S.~Subramanian, K.~Kawakami, and C.~Dyer, ``Neural
  architectures for named entity recognition,'' \emph{arXiv preprint
  arXiv:1603.01360}, 2016.

\bibitem{agrawal2016detecting}
A.~Agrawal, R.~Sahdev, H.~Davoudi, F.~Khonsari, A.~An, and S.~McGrath,
  ``Detecting the magnitude of events from news articles,'' in \emph{Web
  Intelligence (WI), 2016 IEEE/WIC/ACM International Conference on}.\hskip 1em
  plus 0.5em minus 0.4em\relax IEEE, 2016, pp. 177--184.

\bibitem{xu2017local}
M.~Xu, H.~Jiang, and S.~Watcharawittayakul, ``A local detection approach for
  named entity recognition and mention detection,'' in \emph{Proceedings of the
  55th Annual Meeting of the Association for Computational Linguistics (Volume
  1: Long Papers)}, vol.~1, 2017, pp. 1237--1247.

\bibitem{hou2017bocnf}
Y.~Hou, H.~Zhang, and S.~Zhou, ``Bocnf: efficient image matching with bag of
  convnet features for scalable and robust visual place recognition,''
  \emph{Autonomous Robots}, pp. 1--17, 2017.

\bibitem{shakeri2016illumination}
M.~Shakeri and H.~Zhang, ``Illumination invariant representation of natural
  images for visual place recognition,'' in \emph{Intelligent Robots and
  Systems (IROS), 2016 IEEE/RSJ International Conference on}.\hskip 1em plus
  0.5em minus 0.4em\relax IEEE, 2016, pp. 466--472.

\bibitem{zhou2014learning}
B.~Zhou, A.~Lapedriza, J.~Xiao, A.~Torralba, and A.~Oliva, ``Learning deep
  features for scene recognition using places database,'' in \emph{Advances in
  neural information processing systems}, 2014, pp. 487--495.

\bibitem{everingham2010pascal}
M.~Everingham, L.~Van~Gool, C.~K. Williams, J.~Winn, and A.~Zisserman, ``The
  pascal visual object classes (voc) challenge,'' \emph{International journal
  of computer vision}, vol.~88, no.~2, pp. 303--338, 2010.

\bibitem{lin2014microsoft}
T.-Y. Lin, M.~Maire, S.~Belongie, J.~Hays, P.~Perona, D.~Ramanan,
  P.~Doll{\'a}r, and C.~L. Zitnick, ``Microsoft coco: Common objects in
  context,'' in \emph{European conference on computer vision}.\hskip 1em plus
  0.5em minus 0.4em\relax Springer, 2014, pp. 740--755.

\end{thebibliography}
}



\end{document}